\documentclass[letterpaper]{article}

\usepackage[final]{nips_2017}
\PassOptionsToPackage{numbers, square}{natbib}
\setcitestyle{numbers, square}
\usepackage{notoccite}
\usepackage[utf8]{inputenc} % allow utf-8 input
\usepackage[T1]{fontenc}    % use 8-bit T1 fonts
\usepackage{hyperref}       % hyperlinks
\usepackage{url}            % simple URL typesetting
\usepackage{booktabs}       % professional-quality tables
\usepackage{amsfonts}       % blackboard math symbols
\usepackage{nicefrac}       % compact symbols for 1/2, etc.
\usepackage{microtype}      % microtypography
\usepackage{graphicx}
\usepackage{float}
\usepackage{amsmath}
\usepackage{subcaption}
\usepackage{tikz} 
\usetikzlibrary{shapes,arrows} 
\usetikzlibrary{arrows.meta}
\usetikzlibrary{shapes.geometric}
\usepackage{xcolor}
\tikzset{%
  >={Latex[width=2mm,length=2mm]},
            base/.style = {rectangle, rounded corners, draw=black,
                           minimum width=2.0cm, minimum height=1cm,
                           text centered},
}
\title{Phase Transitions in Image Denoising via Sparsely Coding Convolutional Neural Networks}

\author{
  Jacob Carroll\\
  Department of Physics \&  Center for Soft Matter and Biological Physics\\
   Virginia Tech\\
   Blacksburg, VA 24061 \\
   \texttt{jac21934@vt.edu}\\
   \And 
   Nils Carlson \\
   Department of Computer Science and Engineering\\
   New Mexico Tech\\
   Socorro NM, 87801\\
   \And 
   Garrett T. Kenyon\\
   CCS-3, Information Sciences\\
   Los Alamos National Laboratory\\
   Los Alamos, NM, 87545
}

\begin{document}
% \nipsfinalcopy is no longer used

\maketitle

\begin{abstract}
Neural networks are analogous in many ways to spin glasses, systems which are known for their rich set of dynamics and equally complex phase diagrams. We apply well-known techniques in the study of spin glasses to a convolutional sparsely encoding neural network and observe power law finite-size scaling behavior in the sparsity and reconstruction error as the network denoises 32$\times$32 RGB CIFAR-10 images. This finite-size scaling indicates the presence of a continuous phase transition at a critical value of this sparsity. By using the power law scaling relations inherent to finite-size scaling, we can determine the optimal value of sparsity for any network size by tuning the system to the critical point and operate the system at the minimum denoising error.
\end{abstract}

\section{Introduction}
\label{Intro}
Spin glasses and neural networks are very analogous and draw many parallels in their dynamics. Generally, a spin glass is a model of disordered magnetism. The simplest model of a spin glass, the Ising model, is a network of $N$ "spins" $\{\sigma_{i}\}$ which take on the discrete values, connected by a weight matrix $J_{ij} \in \mathbb{R}$ that represents the strength of connection between the spins.  The dynamics of these systems is determined by the values of randomly chosen $J_{ij}$, which are generally time independent \citep{book:Spin_Glasses_Neural_Networks}. 

The similarity of these spin glass systems with neural networks is of interest to us because spin glasses have been a focus of research in statistical physics for the last fifty years, and a large library of machinery and techniques has been developed to deal with them. We would like to apply this machinery to the field of neural networks. 

For this paper we used PetaVision, a high performance neural simulation toolbox \citep{petavision}, to construct sparsely coding convolutional neural networks and examine the relationship between the network's efficiency and sparsity. Interesting behavior in the efficiency of the networks as the sparsity was varied led us to analyze the finite-size scaling of the network, a technique more commonly used in the study of spin glasses, and discovered power law relationships that indicate a continuous (second-order) phase transition is occurring in the networks as sparsity is varied.
\section{Neural network}
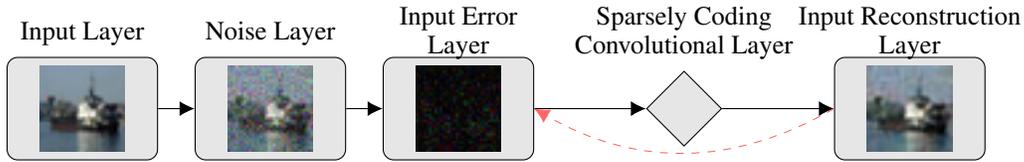
\begin{figure}[h]
\begin{tikzpicture}
\node(Input)[base, fill=black!10] at (0,0){\includegraphics{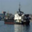}};
\node[above of=Input] {Input Layer};
\node(Noise)[base, fill=black!10] at (2.5,0){\includegraphics{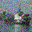}};
\node[above of=Noise] {Noise Layer};
\node(InputErr)[base, fill=black!10] at (5,0){\includegraphics{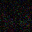}};
\node[above of=InputErr] {\begin{tabular}{c}
Input Error\\
Layer
\end{tabular}};
\node(V1)[diamond,fill=black!10, draw=black,minimum width = 1cm,minimum height=1cm] at (8.0, 0){};
\node[above of=V1] {\begin{tabular}{c}
Sparsely Coding\\
Convolutional Layer
\end{tabular}};
\node(Recon)[base, fill=black!10] at (11,0){\includegraphics{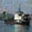}};
\node[above of=Recon] {\begin{tabular}{c}
Input Reconstruction\\
Layer
\end{tabular}};
\draw[->] (Input) -- (Noise);
\draw[->] (Noise) -- (InputErr);
\draw[->] (InputErr) -- (V1);
\draw[->] (V1) -- (Recon);
  \draw[->,style=dashed, draw=red!60] (Recon.west) to [bend left] (InputErr.east);
\end{tikzpicture}
\caption{\label{Denosing_Network}A schematic of the denoising network. The Input Error Layer computes the difference between the Noise Layer and the Input Reconstruction Layer, an alternative implementation of lateral inhibition in LCA \citep{schultz2014replicating}.  Feature learning utilizes a local Hebbian rule to implement stochastic gradient descent.}
\end{figure}

We used two networks in our simulation, both built using PetaVision.
The first network was a sparse auto-encoder network that trained the filter kernels of a convolutional layer using a Locally Competitive Algorithm, as defined by \citet{rozell}, as it attempted to iteratively converge on a sparse representation of different input images. The second network (see Figure \ref{Denosing_Network})  used the same sparsely encoding convolutional layer that was trained by the autoencoder to denoise images that had very high Gaussian noise added to them.

The input for both networks were images from the CIFAR-10 image set \citep{CIFAR-10}.The image set was divided into two parts. The first 50,000 images were used for training the filter kernels of the sparsely coding convolutional layer for different levels of sparsity.  Then, 10,000 additional images had very high Gaussian noise added to them and were denoised by the denoising network for each level of sparsity used in training.

We observed a distinct minimum in the percent reconstruction error of the noisy images as the sparsity of the network was varied that displayed behavior analogous to continuous phase transitions seen in spin glasses (see figure \ref{fig:Error_vs_Fraction}) \citep{book:Spin_Glasses_Neural_Networks,tauber-book}. With this as our motivation we investigated the presence of a phase transition in our system.

\section{Phase transitions and finite-size scaling}
\label{Phase}
A phase of a system is defined as a subspace of the microscopic system parameters where the system's dynamics obey the same macroscale laws and relations everywhere in that subspace. The space of system parameters can have many phases, and the system can transition between them as system control parameters change. The point of transition between two (or more) phases is known as the critical point.

Phase transitions have been subject of significant study in Condensed Matter Physics, and it is well established that the occurrence of a continuous phase transition\footnote{A continuous (second-order) phase transition has a continuous change in the dynamics of the system as it transitions between phases, while first-order phase transitions are discontinuous.} is accompanied by a singularity at the critical point in one or more system parameters when the system is of infinite size \citep{tauber-book}. It is impossible to achieve infinite system sizes computationally, but this theory can be expanded to finite systems where these singularities become truncated and rounded. These minima or maxima that the singularities turn into at finite system sizes follow very specific relations with system size:
\begin{align}
\text{Location of Minima} \sim L^{-1/\nu} \label{eq:Loc}\\
% \end{equation}
% \begin{equation}
\text{Height of Minima} \sim L^{-\gamma/\nu}\label{eq:Height},
\end{align}

where $L$ is the linear system size. This behavior is known as finite-size scaling \citep{tauber-book,Cardy}. The exponents $\nu$ and $\gamma$ two examples of "critical exponents". The critical exponents describe the behavior of the system as it approaches the critical point \citep{tauber-book,Cardy}. Thus we can identify a phase transition in our network by the existence and behavior of minima and maxima in the space of system parameters as we vary the system size, which in our case will be the number of neurons in the convolutional layer. The exponents we record, $\bar{\nu}$ and $\bar{\gamma}$, will be proportional to $\gamma$ and $\nu$  through some effective dimension of our system.
\section{Results}
\label{Results}
The parameters of the system that we are interested in are the fraction of active neurons and the average percent reconstruction error of our noisy images:
\begin{equation}
    P_{err} = \frac{1}{10,000}\sum_{i=1}^{10,000} \frac{\| \mathbf{s}_{i} - \mathbf{\hat{s}}_{i} \|^2}{\|\mathbf{s}_i\|^2}, 
\end{equation}
where $P_{err}$ is the average percent reconstruction error, $\mathbf{s}_{i}$ is the $i^{\text{th}}$ original image before it has Gaussian noise added to it, $\mathbf{\hat{s}}_{i}$ is the $i^{\text{th}}$ reconstruction of the noised image taken from the sparsely coding convolutional layer \citep{petavision,rozell,schultz2014replicating}. 

\begin{figure}[H]
	\centering
	\includegraphics[width=.45\linewidth]{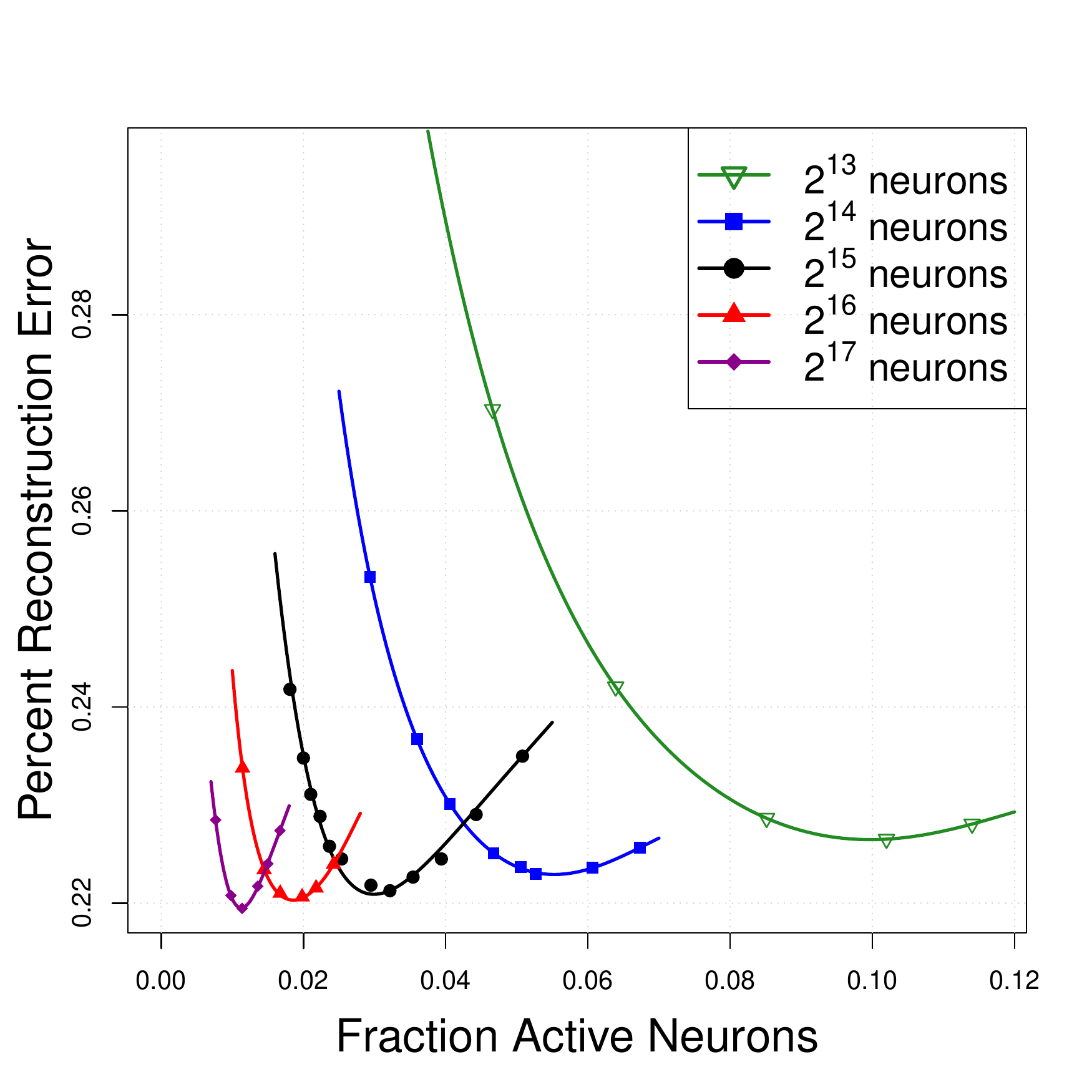}
	\caption{\label{fig:Error_vs_Fraction}Average percent active error vs. fraction active neurons}	
\end{figure}

The fraction of active neurons is controlled by a parameter $\lambda$, as described \citet{rozell}, that behaves monotonically with the sparsity of active neurons and inversely with the fraction of active neurons. Through $\lambda$ we can control the fraction of active neurons and observe how the average percent reconstruction error behaves as the fraction of active neurons is varied. We observed a minimum in $P_{err}$ occur as we varied the fraction of active neurons for many different system sizes. These results are summarized in Figure \ref{fig:Error_vs_Fraction}. 
\begin{figure}[h]
\begin{subfigure}[c]{.4\textwidth}
\includegraphics[width=\textwidth]{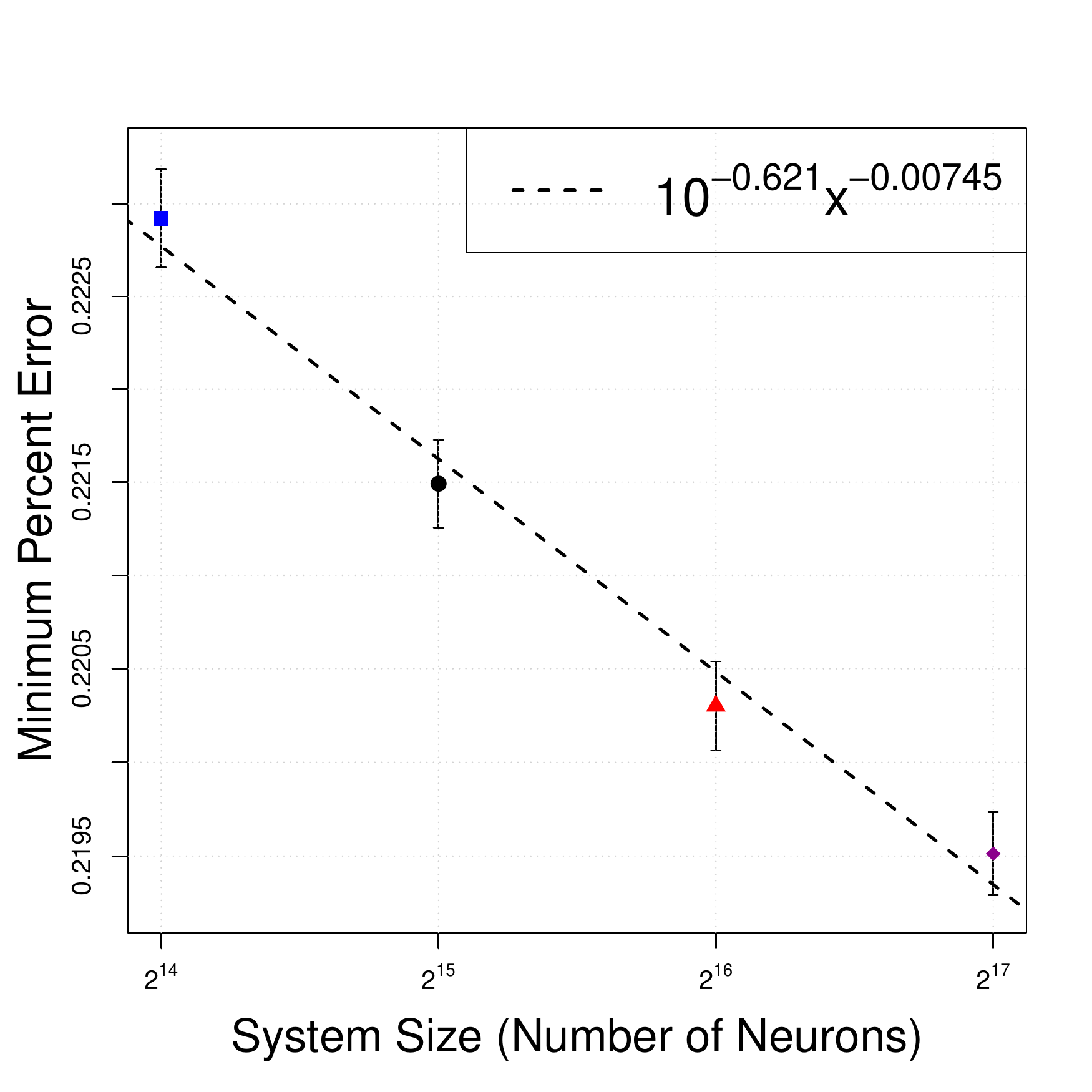}
\caption{Height of minima vs. system size}
\end{subfigure}\hfill%
\begin{subfigure}[c]{.4\textwidth}
\includegraphics[width=\textwidth]{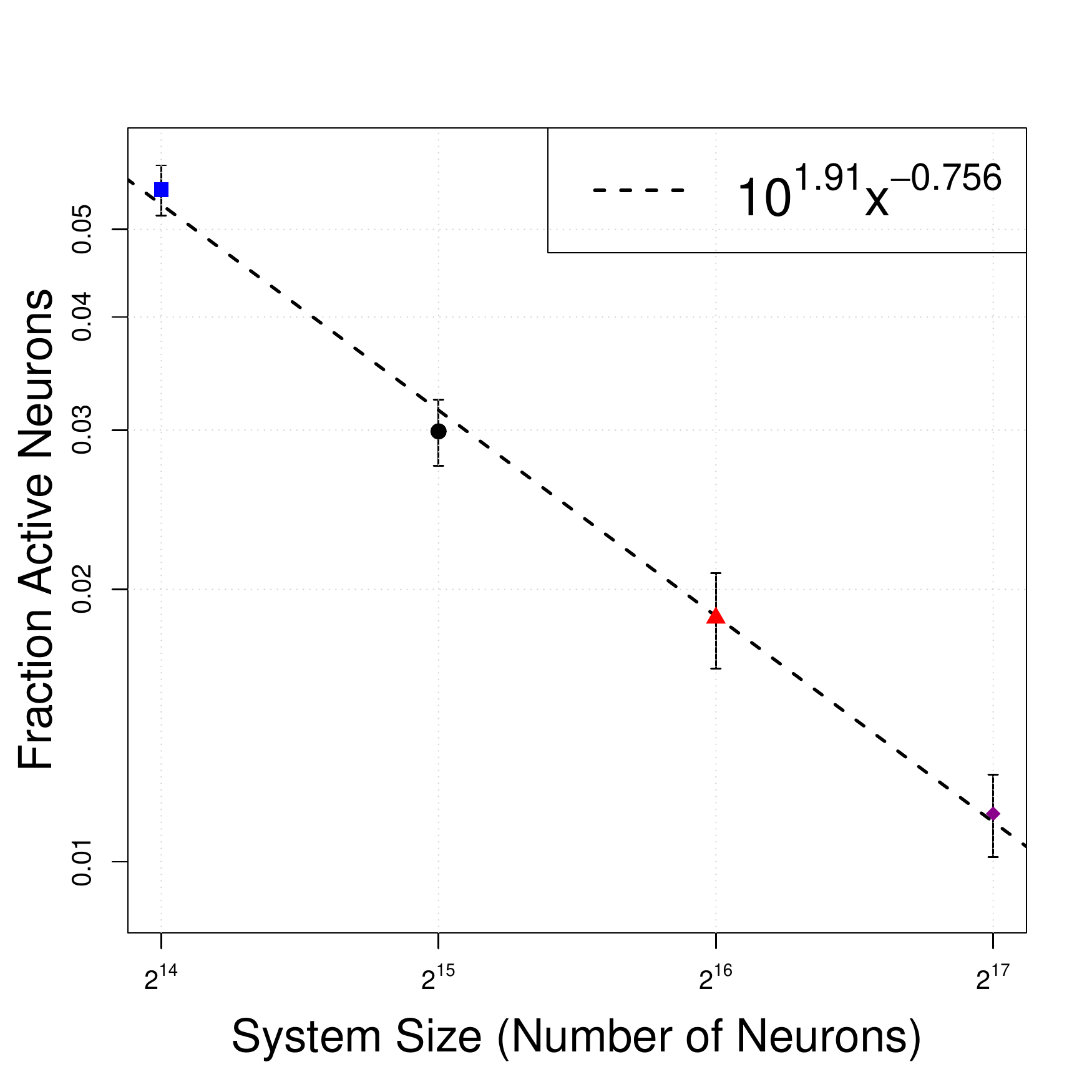}
\caption{Location of minima vs system size}
\end{subfigure}
\caption{\label{fig:results}The power law behavior of the minimum average percent reconstruction error (a), and the fraction of active neurons at that minimum (b). We report $\bar{\nu} =1.32 \pm 0.04$ and $\bar{\gamma} = 0.0099 \pm 0.0095$.}
\end{figure}

We measured the shift in height and location of the minima in $P_{err}$ as the system size was varied, and plot each on a log-log plot (see Figures \ref{fig:results} (a), and \ref{fig:results} (b)). We observe power law behavior in both the location and height of the minima as the system size is varied. This satifies the finite-size scaling requirements as defined in equations \ref{eq:Loc} and \ref{eq:Height}. This finite-size scaling behavior indicates a continuous phase transition is occurring as the sparsity of the network is varied.

\section{Discussion}
The existence of phase transitions in neural networks is not unique to this sparsely coding convolutional system. The auto-associative network proposed by \citet{Hopfield} was shown by \citet{hertz} to display a first-order phase transition in its memory capacity. If the number of patterns recorded by the network exceeds a "critical fraction" of the network size, the output of the network is maximally disordered \citep{hertz}. 

We propose a similar mechanism is responsible for the observed continuous phase transition of our sparsely coding convolution network, where the fraction of active neurons is analogous to the "critical fraction" of learned patterns in the auto-associative network. If our network's fraction of active neurons is too far above the "critical fraction", the network will have the freedom to reconstruct the noise in the image, while if the fraction of active neurons is too low, the network will only reconstruct image components for which it has learned strong priors. These two different regions of dynamics form our "phases". The existence of a phase transition in the average percent reconstruction error of the network  as the fraction of active neurons is varied guarantees the persistence of the power law behavior seen in Figure \ref{fig:results} (b). This power law behavior allows us to predict the optimal fraction of active neurons for any system size, which in turn can be tuned to through the parameter $\lambda$, as described by \citet{rozell}, to ensure that any sparsely coding convolutional network is operating at the optimal level of sparsity. 

The critical behavior of the network allows us to always achieve the minimum denoising error by operating the network at this critical value of sparsity.
\subsubsection*{Acknowledgments}
We gladly acknowledge helpful discussions with Uwe C. T\"auber.

This work was supported by the Los Alamos National Laboratory under contract DE-AC52-06NA25396.

Computations were performed using the Darwin Computational Cluster at Los Alamos National Laboratory.
%\section*{References}
\medskip
\small
\bibliography{ref}

\begin{thebibliography}{9}
\providecommand{\natexlab}[1]{#1}
\providecommand{\url}[1]{\texttt{#1}}
\expandafter\ifx\csname urlstyle\endcsname\relax
  \providecommand{\doi}[1]{doi: #1}\else
  \providecommand{\doi}{doi: \begingroup \urlstyle{rm}\Url}\fi

\bibitem[pet()]{petavision}
Petavision.
\newblock URL \url{github.com/PetaVision/OpenPV}.

\bibitem[Cardy(1996)]{Cardy}
J.~Cardy.
\newblock \emph{Scaling and renormalization in statistical physics}.
\newblock Cambridge University Press, 1996.
\newblock ISBN 0521499593.

\bibitem[Dotsenko(1995)]{book:Spin_Glasses_Neural_Networks}
V.~Dotsenko.
\newblock \emph{An introduction to the theory of spin glasses and neural
  networks}.
\newblock World Scientific Lecture Notes in Physics. World Scientific
  Publishing Company, 1995.
\newblock ISBN 9810218737.

\bibitem[Hertz et~al.(1991)Hertz, Krogh, and Palmer]{hertz}
J.~A. Hertz, A.~S. Krogh, and R.~G. Palmer.
\newblock \emph{Introduction to the theory of neural computation}.
\newblock Addison-Wesley Publishing Company, 1991.
\newblock ISBN 0201515601.

\bibitem[Hopfield(1982)]{Hopfield}
J.~J. Hopfield.
\newblock Neural networks and physical systems with emergent collective
  computational abilities.
\newblock \emph{Proceedings of the National Academy of Sciences}, 79\penalty0
  (8):\penalty0 2554--2558, 1982.
\newblock URL \url{http://www.pnas.org/content/79/8/2554.abstract}.

\bibitem[Krizhevsky(2009)]{CIFAR-10}
A.~Krizhevsky.
\newblock Learning multiple layers of features from tiny images.
\newblock 2009.

\bibitem[Rozell et~al.(2008)Rozell, Johnson, Baraniuk, and Olshausen]{rozell}
C.~J. Rozell, D.~H. Johnson, R.~G. Baraniuk, and B.~A. Olshausen.
\newblock Sparse coding via thresholding and local competition in neural
  circuits.
\newblock \emph{Neural Computation}, 20:\penalty0 2526--2563, 2008.

\bibitem[Schultz et~al.(2014)Schultz, Paiton, Lu, and
  Kenyon]{schultz2014replicating}
P.~F. Schultz, D.~M. Paiton, W.~Lu, and G.~T. Kenyon.
\newblock Replicating kernels with a short stride allows sparse reconstructions
  with fewer independent kernels.
\newblock \emph{arXiv preprint arXiv:1406.4205}, 2014.

\bibitem[T\"{a}uber(2014)]{tauber-book}
U.~C. T\"{a}uber.
\newblock \emph{Critical dynamics: a field theory approach to equilibrium and
  non-equilibrium scaling behavior}.
\newblock Cambridge University Press, 2014.
\newblock ISBN 9780521842235.

\end{thebibliography}
% \printbibliography
\end{document}